\documentclass{llncs}
\usepackage{isolatin1}
\usepackage{epsf,epsfig}
\usepackage{graphics}

\newcommand{\revise}[2]{#2}

\title{Evolving Symbolic Controllers} 
\author{Nicolas Godzik$^1$ \and Marc Schoenauer$^1$ \and Michèle
  Sebag$^2$}

\institute{
Projet Fractales, INRIA Rocquencourt, France \and LRI, Université Paris-Sud, France\\~\\
Published in G. Raidl et al., eds, {\em Applications of Evolutionary Computing}, \\
pp 638-650, LNCS 2611, Springer Verlag, 2003.\\~\\
}
\date{}
\begin{document}
\maketitle

\vspace{-0.8cm}
\begin{abstract}
The idea of {\em symbolic controllers} tries to bridge the gap
between the top-down manual design of the controller architecture, as
advocated in Brooks' subsumption architecture, and the bottom-up
designer-free approach that is now standard within the Evolutionary Robotics
community. The designer provides a set of elementary behavior, and
evolution is given the goal of assembling them to solve complex
tasks. Two experiments are presented, demonstrating the efficiency and
showing the recursiveness of this approach. In particular, the sensitivity
with respect to the proposed elementary behaviors, and the robustness
w.r.t. generalization of the resulting controllers are studied in detail. 
\end{abstract}

\section{Introduction}

There are two main trends in autonomous robotics. 
\revise{The first one,
advocated by R. Brooks \cite{Brooks-subsumption86}, is a top-down
deterministic approach: the tasks of the robot are manually decomposed
into a hierarchy of independent sub-tasks, resulting in the  the
so-called {\em  subsumption architecture}.}
{There are two main trends in autonomous robotics. The first one,
advocated by R. Brooks \cite{Brooks-subsumption86}, is a
human-specified deterministic approach: the tasks of the robot are
manually decomposed 
into a hierarchy of independent sub-tasks, resulting in the  the
so-called {\em  subsumption architecture}.}

\revise{
On the other hand, evolutionary robotics (see
e.g. \cite{Nolfi:Floreano:Book00}), can be considered as a bottom-up 
approach: black-box  
controllers, mapping the sensors to the actuators, are optimized using
the Darwinian paradigm of Evolutionary Computation; the programmer
only designs the fitness function.}
{On the other hand, evolutionary robotics (see
e.g. \cite{Nolfi:Floreano:Book00}), is generally viewed  as a pure
black-box approach: some 
controllers, mapping the sensors to the actuators, are optimized using
the Darwinian paradigm of Evolutionary Computation; the programmer
only designs the fitness function.}

However, the scaling issue remains critical for both approaches,
though for different reasons.
The efficiency of the \revise{top-down}{human-designed} approach is
limited by the human 
factor: it is very difficult to decompose complex tasks into the
subsumption architecture.
On the other hand, the answer of evolutionary robotics to the
complexity challenge is very often to come up with an ad hoc (sequence
of) specific fitness function(s). The difficulty is
transferred from the internal architecture design to some external action on
the environment. Moreover, the black-box approach makes it extremely
difficult to understand the results, be they successes or failures,
hence forbidding any  capitalization of past expertise for further re-use.
This issue is discussed in more details in section \ref{Discu}.

The approach proposed in this work tries to find some compromise
between the two extremes mentioned above. It is based on the
following remarks: First, scaling is one of the most critical issues
in autonomous robotics. Hence, the {\em same mechanism} should be used
all along the complexity path, leading from primitive tasks to simple
tasks, and from simple tasks to more complex behaviors. 

One reason for the lack of intelligibility is that the language of the
controllers consists in low-level orders to the robot actuators
(e.g. speeds of the right and left motors). Using instead hand-coded basic
behaviors (e.g. forward, turn left or turn right) as proposed in
section \ref{symbolic} should allow one to
better understand the relationship between the controller outputs and
the resulting behavior. Moreover, the same
approach will allow to recursively build higher level behaviors from
those evolved simple behaviors, thus solving more and more complex
problems. 

Reported experiments tackle both aspects of the above approach. After
describing the \revise{simulated) }{~}
experimental setup in section \ref{setup}, 
simple behaviors are evolved based on basic primitives (section
\ref{simple}). Then a more complex task is solved using the results
of the first step (section \ref{complexe}). 
Those results are validated by comparison to the
pure black-box evolution, and important issues like the sensitivity
w.r.t. the available behaviors, and the robustness w.r.t
generalization are discussed. The paper ends by revisiting the discussion
in the light of those  results.

\section{State of the art} \label{Discu}
The trends in autonomous robotics can be discussed in the light of the
``innate vs acquired'' cognitive aspects -- though at the time scale
of evolution.
\revise{Along those lines}{From that point of view}, 
Brooks'subsumption architecture 
is extreme on the
``innate'' side: The robots are given all necessary skills by their
designer, from basic behaviors to the way to combine them.
Complex behaviors so build on some ``instinctive'' 
predefined simple behaviors. Possible choices lie in a very
constrained space, guaranteeing good performances for very specific
tasks, but that does not scale up very well: Brooks'initial goal
was to reach the intelligence of insects \cite{Brooks:howto91}.

\revise{
Another position that lies close to Brooks'one is that of 
action-selection through reinforcement learning \cite{Kaelbling}. 
The actions are pre-defined,
and only their selection is learned -- on-line. }
{Along similar lines are several computational models of
  action-selection (e.g. Spreading Activation Networks
  \cite{Maes:IJCAI89}, reinforcement learning
  \cite{Kaelbling,Humphrys:PhD97}, \ldots).}
Such \revise{approach has}{approaches have} two main
weaknesses. The first one is that such
architecture is biologically \revise{arguable}{questionable} -- 
but do we really care here? 
The second weakness is concerned with the autonomy issue. 
Indeed, replacing low level reflexes by
decisions about high level behaviors (use this or that behavior now)
might be beneficial. However, how to program the {\em interactions} of such
reflexes in an open world amounts to solve the {\em exploration vs 
  exploitation} dilemma -- and both Game Theory and Evolutionary
Computation have underlined the difficulty of answering this question.

At the other extreme of the innate/acquired spectrum is the
evolutionary robotics credo: any a priori bias from the designer can be
harmful. Such position is also defended by Bentley
\cite{Bentley:book:99} in the domain of optimal design, where 
it has been reinforced by some very unexpected excellent solutions
that arose from evolutionary design processes.
In the Evolutionary Robotics area,  this idea has been sustained by the
recent revisit  by 
Tuci, Harvey and Quinn \cite{Tuci:Harvey:SAB02} of an experiment initially
proposed  by Yamauchi and Beer \cite{Yamauchi:Beer:SAB94}:
depending on some random
variable, the robot should behave differently (i.e. go toward the
light, or away from it). The robot must hence learn from the
first epoch the state of that random variable, and act accordingly in
the following epochs.

The controller
architecture is designed manually in the original experience, whereas
evolution has complete freedom in its recent remake 
\cite{Tuci:Harvey:SAB02}. Moreover, Tuci et al. use no explicit
reinforcement.
Nevertheless, the results obtained by this recent approach are much better
than the original results - and the
authors claim that the reason for that lies in their complete black-box
approach.

However, whereas the designers did decide to use a specifically
designed modular 
architecture in the first experiment, the second experience required a
careful design of the fitness function
(for instance, though the reward lies under the light only half of the
time, going toward the light has to be rewarded more than fleeing away
to break the symmetry). So be it at the ``innate'' or ``acquired''
level, human intervention is required, and must act at some very
high level of subtlety.

Going beyond this virtual ``innate/acquired'' debate, an intermediate
issue  would be to be able to evolve complex controllers that could
benefit from human knowledge but that would not require high level of
intervention with respect to the complexity of the target task.

Such an approach is what is proposed here: the designer is supposed to help
the evolution of complex controllers by simply seeding the process
with some 
simple behaviors -- hand-coded or evolved -- letting evolution
arrange those building blocks together. An important side-effect is
that the designer will hopefully be able to better {\em understand}
the results of an 
experiment, because of the greater intelligibility of the controllers.
It then becomes easier to manually optimize the experimental protocol,
e.g. to gradually refine the fitness in order to solve some very
complex problems.

\section{Symbolic controllers}\label{symbolic}

\subsection{Rationale}
\label{highLevel}
The proposed approach pertains to Evolutionary Robotics 
\cite{Nolfi:Floreano:Book00}. Its originality lies in the
representation space of the controllers, i.e. the search space of the
evolutionary process. One of the main goals is that the results will be
intelligible enough to allow an easy interpretation of the results,
thus easing the whole design process.

A frequent approach in Evolutionary Robotics is to use Neural Networks
as controllers (feedforward or recurrent, discrete or continuous). The
inputs of the controllers are the sensors (Infra-red, camera, \ldots),
plus eventually some reward ``sensor'', either direct
\cite{Yamauchi:Beer:SAB94} or indirect \cite{Tuci:Harvey:SAB02}. 

The outputs of the controllers are the actuators of the robot
(e.g. speeds of the left and right motors for a Khepera robot).
The resulting controller is hence comparable to a program in machine
language, thus difficult to interpret. To overcome this difficulty, we
propose to use higher level outputs, namely involving four possible
actions: {\em Forward, Right, Left} and {\em Backward}. In order to
allow some flexibility, each one of these symbolic actions should be
tunable by some continuous parameter (e.g. speed of forward
displacement, or turning angle for left and right actions). 

The proposed {\em symbolic controller}  has eight outputs with
values in $[0,1]$: the first four outputs are used to specify which
action will be executed, namely action $i$, with  
 $i = Argmax (output(j), j = 1..4)$. Output  $i+4$ then gives the
 associated  parameter.
From the given action and the associated parameter, the values of the
commands for the actuators are computed by some simple hard-coded
program.

\subsection{Discussion}\label{plateau}

Using some high level representation language for the
controller impacts on both the size of the search space, and the
possible modularity of the controller.

At first sight, it seems that the size of the search space is
increased, as a symbolic controller has more outputs than a classical
controller. However, at the end of the day, the high level actions are
folded into the two motor commands. On the other hand, using a
symbolic controller can be viewed as adding some constraints on the
search space, hence reducing the size of the part of the
search space actually explored. The argument here is similar to the
one used in statistical learning \cite{Vapnik98}, where rewriting
the learning problem into a very high dimensional space actually makes
it simpler.
Moreover, the fitness landscape of the
space that is searched by a symbolic controller has many neutral
plateaus, as only the highest value of the first outputs is used --
and neutrality can be beneficial to escape local optima
\revise{\cite{Shipman:neutrality:AL2000}}{\cite{Harvey:PhD93}}. 

On the other hand, the high level primitives of symbolic controllers
make room for modularity.
And according to Dawkins \cite{Dawkins88}, the probability to build a
working complex system by a randomized process increases with the
degree of modularity.It should be noted that this principle is already
used in Evolutionary Robotics, for instance to control the robot gripper~: 
the outputs of the controllers used in \cite{Nolfi:Floreano:Book00} are
high-level actions (e.g. {\em GRAB, RAISE GRIPPER, \ldots}), and not the
commands of the gripper motors.

Finally, there are some similarities between the symbolic controller
approach and reinforcement learning\revise{\cite{Sutton}.
The main difference
is that symbolic controller do no have to quantize the action space,
whereas reinforcement learning is limited by the number of possible
actions.}
{. Standard reinforcement learning \cite{Sutton,Kaelbling}
aims at finding an optimal
  policy. This requires an intensive exploration of the search
  space. In contrast, evolutionary approaches sacrifices optimality
  toward satisfactory timely satisfying solutions. More recent
  developments \cite{Millan:MLJ2002}, closer to our approach, handle
  continuous state/action spaces, but rely on the specification 
of some relevant initial policy involving manually designed ``reflexes''. 
}

\section{Experimental setup}
\label{setup}
\revise{
Only simulated  experiments have been performed so far. The Khepera
simulator that has been used,  {\tt EOBot}, was developed by the
first author from the  {\tt EvoRobot} software provided by S. Nolfi and
D. Floreano \cite{Nolfi:Floreano:Book00}. 
}
{
Initial experiments have been performed using the Khepera simulator
  {\tt EOBot}, that was developed by the
first author from the  {\tt EvoRobot} software provided by S. Nolfi and
D. Floreano \cite{Nolfi:Floreano:Book00}.
} 
{\tt EvoRobot} was ported
on Linux platform using {\tt OpenGL} graphical library, and interfaced
with the {\tt EO} library \cite{EO:EA01}. It is hence now possible to use
all features of {\tt EO} in the context of Evolutionary Robotics, 
\revise{
from alternate evolution engines (e.g. other selection and replacement
procedures, multi-objective optimization, co-evolution, \ldots) to
other representations and paradigms (e.g. Evolution Strategies,
Genetic Programming, \ldots)
}
{
e.g. other selection and replacement
procedures, multi-objective optimization, and even 
other paradigms like Evolution Strategies and Genetic Programming.
}
However, all experiments presented here use as controllers Neural
Networks (NNs) with fixed topology. The genotype is hence the vector of
the (real-valued) weights of the NN.
Those weights evolve in  $[-1,1]$ (unless otherwise mentioned), 
using a  $(30,150)$-Evolution
Strategy with intermediate crossover and self-adaptive Gaussian mutation
\cite{Schwefel}:  
Each one of 30 parents gives birth to 5 offspring, and the best 30 of
the 150 offspring become the parents for next generation. 
All experiments  run for 250 generations,
requiring about 1h to 3h depending on the experiment.

All results shown in the following are statistics based on at least 10
independent runs. One fitness evaluation is made of 10 {\em epochs},
and each epoch 
lasts from 150 to 1000 time steps (depending on the experiment),
starting from a  randomly chosen initial position.

\section{Learning a simple behavior: obstacle avoidance}\label{simple}

\subsection{Description}
The symbolic controller (SC) with 8 outputs, described in section
\ref{highLevel}, is compared 
to the classical controller (CC) the outputs of which are the speeds
of both motors. 
Both controllers have 8 inputs, namely the IR sensors of the Khepera
robot in active mode (i.e. detecting the obstacles).

Only Multi-Layer Perceptrons (MLP) were considered for this simple
task. Some preliminary  experiments with only two layers (i.e. without
hidden neurons) demonstrated better results for the  Symbolic
Controllers than for the Classical Controller. Increasing the number
of hidden neurons increased the performance of both types of
controllers. Finally, to make the comparison ``fair'', the following
\revise{near-optimal}{~}
architectures were used: 14 hidden neurons for the SC, and 20 hidden
neurons for the CC, resulting in roughly the same number of weights
(216 vs 221).

The fitness function\revise{, inspired by that of
  \cite{Nolfi}}{\cite{Nolfi}}, is defined as
$ \sum_{epoch} \sum_t |V(t)| (1 - \sqrt{\delta V(t)}) $, 
where $V(t)$ is the average speed of the robot at time $t$, and 
$\delta V(t)$ the absolute value of the difference between the speeds
of the left and right motors\revise{(in order to avoid a fast rotation of the
robot that would maximize both motor speeds).}{.} The difference with the
original fitness function is the lack of IR sensor values in the
fitness: the obstacle avoidance behavior is here implicit, as an epoch
immediately ends whenever the robot hits an obstacle. The arena is
similar to that in \cite{Nolfi}.

\subsection{Results}
\label{loophole}
The first results for the SC were surprising: half of the runs,
even without hidden neurons, 
find a loophole in the fitness function: due to the absence of inertia
in the simulator, an optimal behavior is obtained by a rapid 
succession of {\em FORWARD - BACKWARD} movements at maximum speed -
obviously avoiding all obstacles! 

A degraded SC that has no {\em BACKWARD} action cannot take advantage
of this bug. Interestingly, classical controllers only discover this
trick when provided with more than 20 hidden neurons {\bf and} if
the weights are searched in a larger interval (e.g. $[-10,10]$).

Nevertheless, in order to definitely avoid this loophole, the fitness
is modified in such a way that it increases only when the robot moves
forward (sum of both motor speeds positive)\footnote{
Further work will introduce inertia in the simulator, thus avoiding
this trap -- and possibly many others.}.

This modification does not alter the ranking of the controllers: the
Symbolic Controller still outperforms the Classical
Controller. This advantage somehow vanishes when more hidden neurons
are added (see Table 1), but the results of the SC exhibit a
much smaller variance.\\

{\small
\begin{tabular}{ccc}
    \begin{tabular}{|l|c|c|}
\hline
Architecture & CC & CS \\
\hline
8-2 / 8-6 & 861 $\pm$ 105 & 1030 $\pm$ 43 \\
8-8-2 / 8-8-6 & 1042 $\pm$ 100 & 1094 $\pm$ 55 \\
8-20-2 / 8-14-6 & 1132 $\pm$ 55 & 1197 $\pm$ 16 \\
8-20-2 / 8-14-6$^*$ & 1220 $\pm$ 41 & 1315 $\pm$ 6 \\
\hline 
    \end{tabular}
& ~ &
\parbox{6cm}{
{{\bf Table 1.} 
{\em Averages and standard deviations for 10 independent runs for
  the obstacle avoidance experiment. 
$^*$ this experiment was performed in a more
constrained environment.}
}
}
\end{tabular}
}




\section{Evolution  of a complex behavior}\label{complexe}

\subsection{Description}
The target behavior is derived the homing experiment first proposed in 
\cite{Floreano:Mondada:TSMC94}, combining exploration of the
environment with energy management.  The robot is equipped with an
accumulator. The robot completely consumes the accumulator energy in
285 times steps. A specific recharge area is signaled by a light in
the arena. There are no obstacles in the arena, and the position of
the recharge area is randomly assigned at each epoch. 

The fitness is increased proportionally to the forward speed of the
robot (as described in section \ref{loophole}), but only when
the robot is not in the recharge area.

In the original experiment \cite{Floreano:Mondada:TSMC94},
the accumulator was instantly recharged when the
robot entered the recharge area. We suppose here that the recharge is
proportional to the time spent in the recharge area (a full recharge
takes 100 times steps). Moreover, the recharge area is not directly 
``visible''
for the robot, whereas it was signaled by a black ground that the
robot could detect with a sensor in \cite{Floreano:Mondada:TSMC94}.
These modifications increase the complexity of the task.

\subsection{Supervisor architectures}
\label{description}
The {\em supervisor} architecture is a hierarchical
controller that decide at each time step which one of the basic
behaviors it supervises will be executed. Its number of outputs is the
number of available basic behaviors, namely:

\begin{itemize}
\item {\em Obstacle avoidance}. This behavior is evolved as described
  in section \ref{loophole}; 
\item {\em Light following}. The fitness used to evolve this behavior
  is the number of times it reaches the light during 10 epoch (no
  energy involved);
\item {\em Stop}. This behavior is evolved to minimize the speed of
  the center of mass of the robot. Note that a very small number of
  generations is needed to get the perfect behavior, but that all
  evolved {\em Stop} behaviors in fact rotate rapidly with inverse speeds on
  both motors.

\item {\em Area sweeping}. The arena is divided in small squares, and
  the fitness is the number of squares that were visited by the robot
  during 10 epoch
\footnote{This performance is external, i.e. it could not be computed
  autonomously by the robot. But the resulting controller only uses
  internal inputs.}.
\end{itemize}

Two types of supervisors have been tested: the Classical Supervisor
(CS), using Classical Controllers as basic behaviors, and the Symbolic
Supervisor (SS), that uses symbolic controllers (see section
\ref{symbolic}) as basic behaviors.
The NN implementing the supervision are are Elman networks 
\footnote{Elman recurrent neural networks are 3 layers MLP 
in which all neurons  of the hidden layer are totally
connected with one another.} with 5 hidden neurons.

Baseline experiments were also performed using {\em direct
  controllers} with the same Elman architecture - either Classical, or
Symbolic (see section \ref{symbolic}).

All supervisors and direct controllers have 17 inputs: the 8 IR sensors in
active mode for obstacle 
detection, the 8 IR sensors in passive mode for ambient light
detection, and the accumulator level.

The number of outputs is 2 for the classical controller, the speeds of
the motors, 6 for the symbolic controller using the three hard-coded
behaviors {\em FORWARD, LEFT, RIGHT} (see section \ref{symbolic}), and
4 for both supervisors (Classical and Symbolic) 
that use the evolved basic behaviors {\em
  obstacle avoidance, light following, area sweeping} and {\em stop}.
The {\em obstacle avoidance} behaviors that are used are the best
results obtained in the experiments of section \ref{simple}. Similar
experiments (i.e. with the same architectures) were run for the {\em
  area sweeper} and the best results of 10 runs were chosen. For the
simpler {\em light following} and {\em stop}, 2-layers networks were
sufficient to get a perfect fitness. 

\begin{center}
\begin{figure}[htbp]
\begin{tabular}{cc}
    \includegraphics[width=6cm]{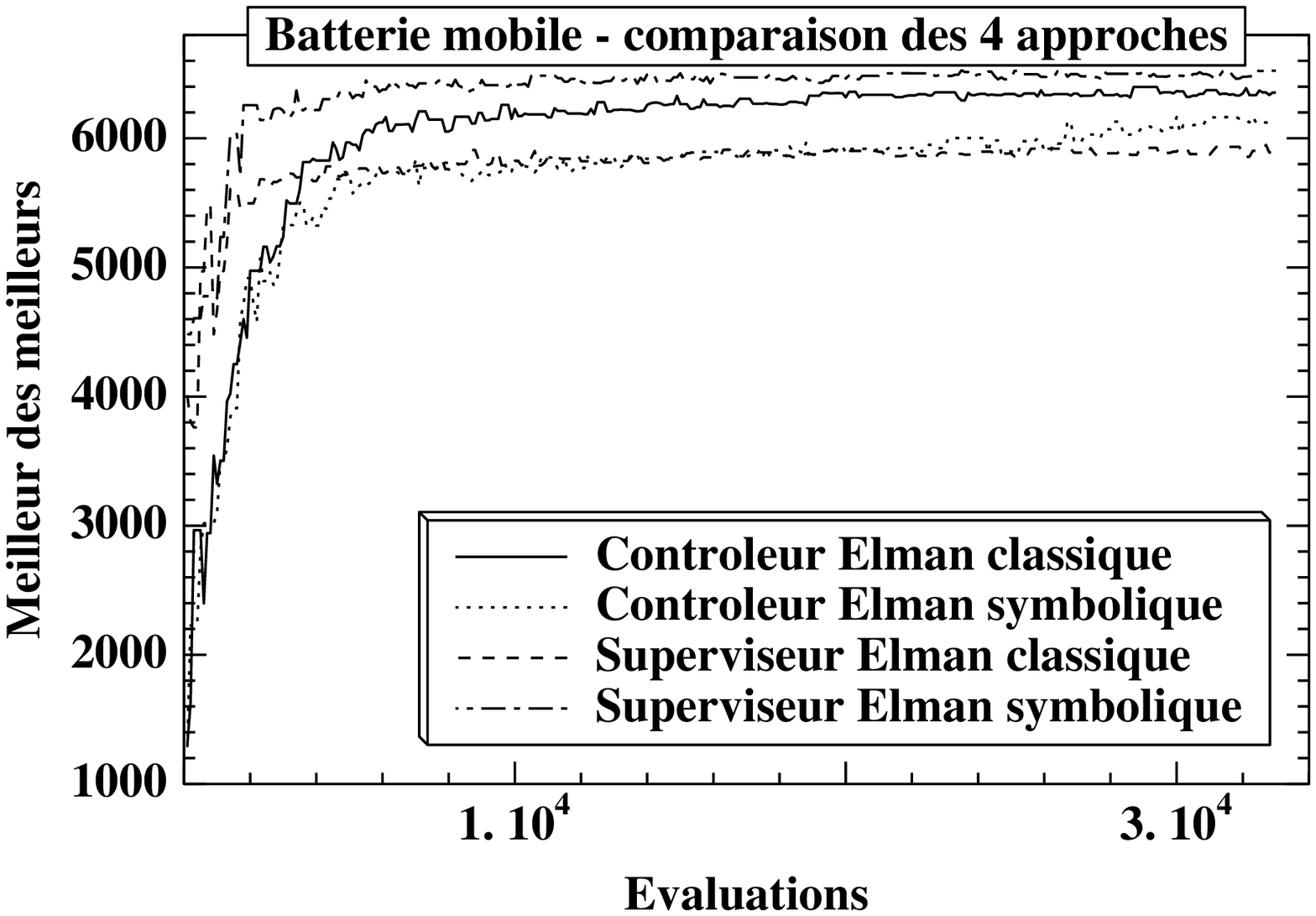}
&
    \includegraphics[width=6cm]{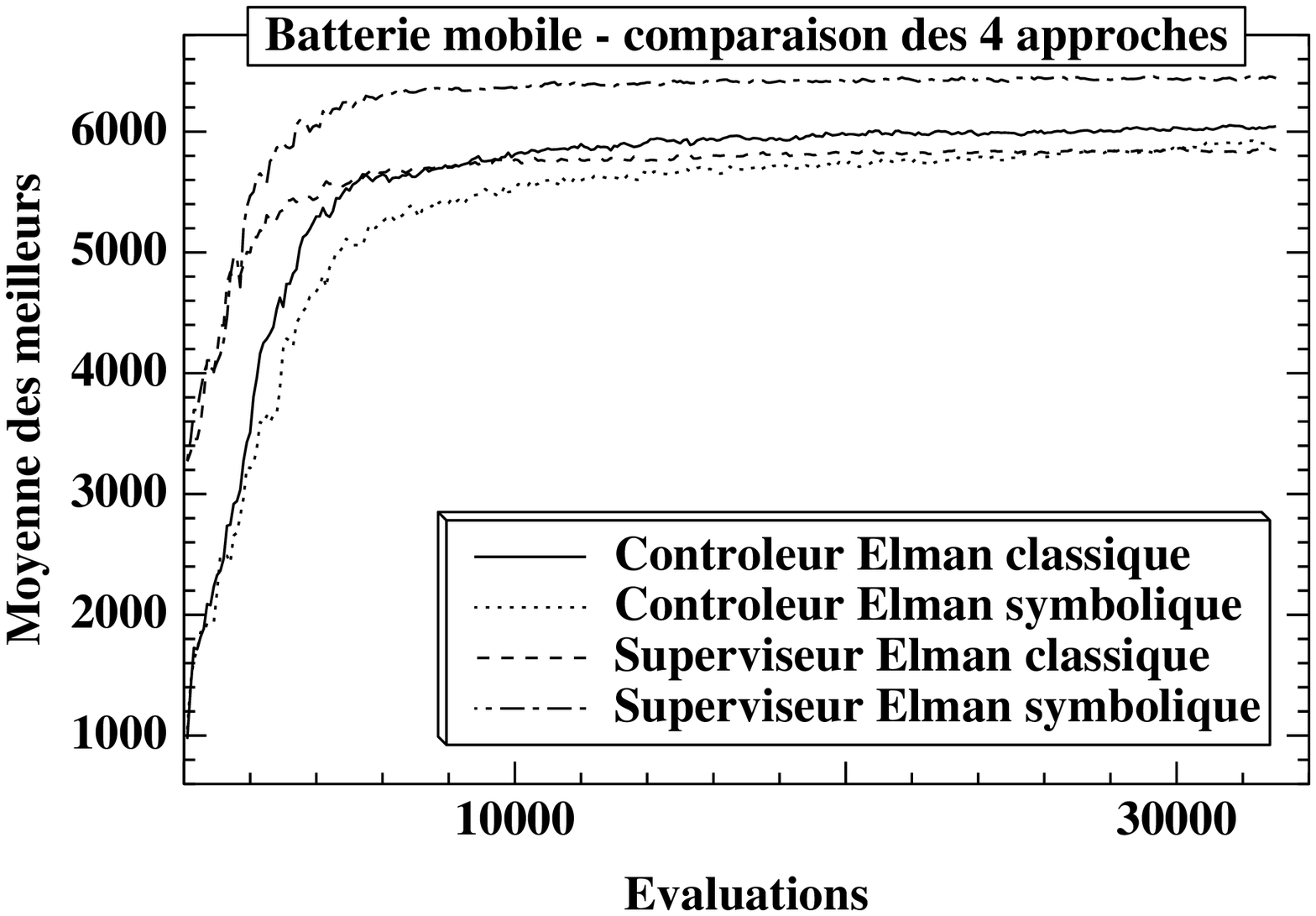}
\end{tabular}

    \caption{Maximum (left) and average (right) of best fitness in the
      population  for the energy  experiments.}
    \label{bestavg4}
\end{figure}
\end{center}

\vspace{-1cm}

\subsection{Results}
\label{complexresults}
The statistics over 10 independent runs can be seen on Figure \ref{bestavg4}.
Three criteria can be used to compare the performances of the 4
architectures: the best overall performance, the variance of the
results, and how rapidly good performances are obtained. The
sensitivity and 
generalization abilities of the resulting controllers are 
important criteria that require additional experiments (sections
\ref{sensitivity},  \ref{generalization}). 

The best overall performance are obtained by the SS (Symbolic
Supervisor) architecture. Moreover, it exhibits a very low variance
(average best fitness is $6442 \pm 28$). Note that overpassing a
fitness of $6400$ means that the resulting behavior could go on for
ever, almost optimally storing fitness between the recharge
phases). 

Next best architecture is the CC (Classical Controller). But whereas
its best overall fitness is only slightly less that that of the SS, the
variance is 10 times larger (average best fitness is $6044 \pm
316$, with best at $6354$). The difference is statistically
significant with 95\% confidence using the Student T-test.

The SC (Symbolic Controller) and CS (Classical Supervisors) come last,
with respective average best fitness of $5902 \pm 122$ and $5845 \pm
27$. \\

Some additional comments can be made about those results. First, both
supervisors architectures exhibit a very good best fitness ($\approx
3200$) in the initial population: such fitness is in fact obtained
when the supervisors only use the obstacle avoidance -- they score
maximum fitness until their accumulator level goes to 0. Of course,
the direct controller architectures require some time to reach
the same state (more than 2000 evaluations).

Second, the variance for both supervisor architectures is very
low. Moreover, it seems that 
this low variance is not only true at the performance level, but also
at the behavior level: whereas all
symbolic supervisors do explore the environment until their energy
level becomes dangerously low, and then head toward the light and {\bf
  stay} in the recharge area until their energy level is maximal
again, most (but not all) of the direct controller architectures seem
to simply stay close to the recharge area, entering it randomly.

One last critical issue is the low performance of the Symbolic
Controller. A possible explanation is the existence of the neutrality
plateaus discussed in section \ref{plateau}:  though those plateaus
help escaping local minima, they also slow down the learning
process. Also it appears clearly on Figure \ref{bestavg4}-left that
the SC architecture is the only one that seems to steadily increase
its best fitness 
until the very end of the runs. Hence, the experiment was carried on
for another 250 generations, and indeed the SC architecture did
continue to improve (over a fitness level of $6200$) -- while all
other architectures simply stagnates.

\label{analysis}
The evolved behaviors have been further examined. Figure
\ref{behaviorcalls}-left shows a typical plot of the number of calls
of each basic behaviors  by the best
evolved Symbolic Supervisor during one fitness evaluation. 
First, it appears
that both supervisors architectures mainly use the {\em obstacle
  avoidance} behavior, never use the {\em area sweeping}, and, more
surprisingly, almost never use the {\em light following}: when they see
the light, they turn using the {\em stop} behavior (that consists in
fast rotation), and then go to the
light using the {\em obstacle avoidance}. However, once on the
recharge area, they use the {\em stop} until the energy level is
back over 90\%.

 Investigating deeper, it appears that the speeds of the {\em light
  following} and {\em area sweeper} are lower than that
 of the {\em obstacle avoidance} -- and speed is crucial in this
experiment. Further experiments will have to modify the speeds of all
behaviors to see if it makes any difference.
However, this also demonstrates that the supervisor can discard some
behavior that proves under-optimal or useless.

\subsection{Sensitivity Analysis}
\label{sensitivity}

One critical issue of the proposed approach is how to ensure that the
``right'' behavior will be available to the supervisor. A possible
solution is to propose a large choice -- but will the supervisor be
able to retain only the useful ones?
In order to assess this, the same energy experiment was repeated but
many useless, or even harmful, behaviors were added to the 4 basic
behaviors\revise{ listed in section \ref{description}.}{.}

\begin{center}
\begin{figure}[htbp]
\begin{tabular}{cc}
    \includegraphics[width=6cm]{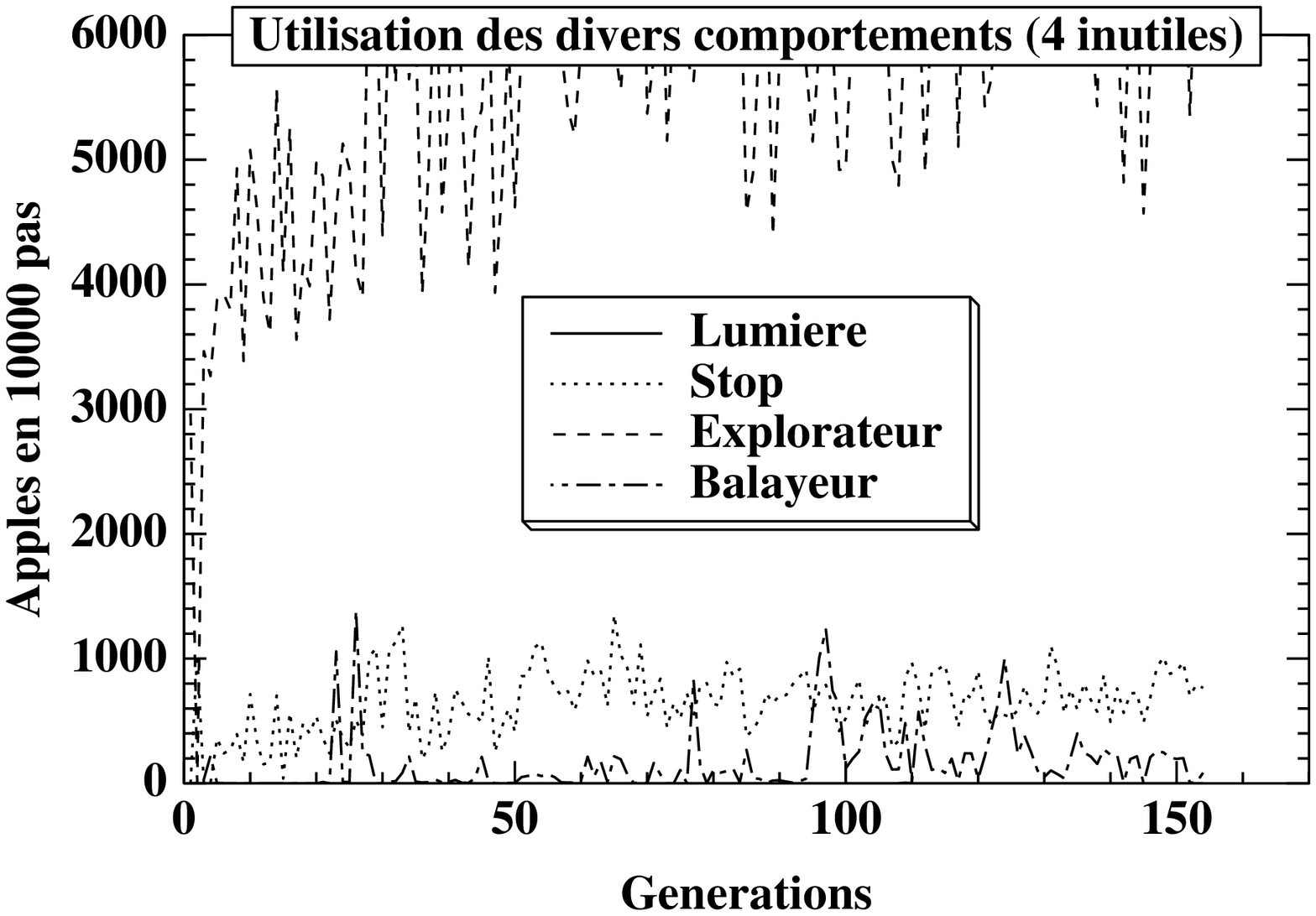}
&
    \includegraphics[width=6cm]{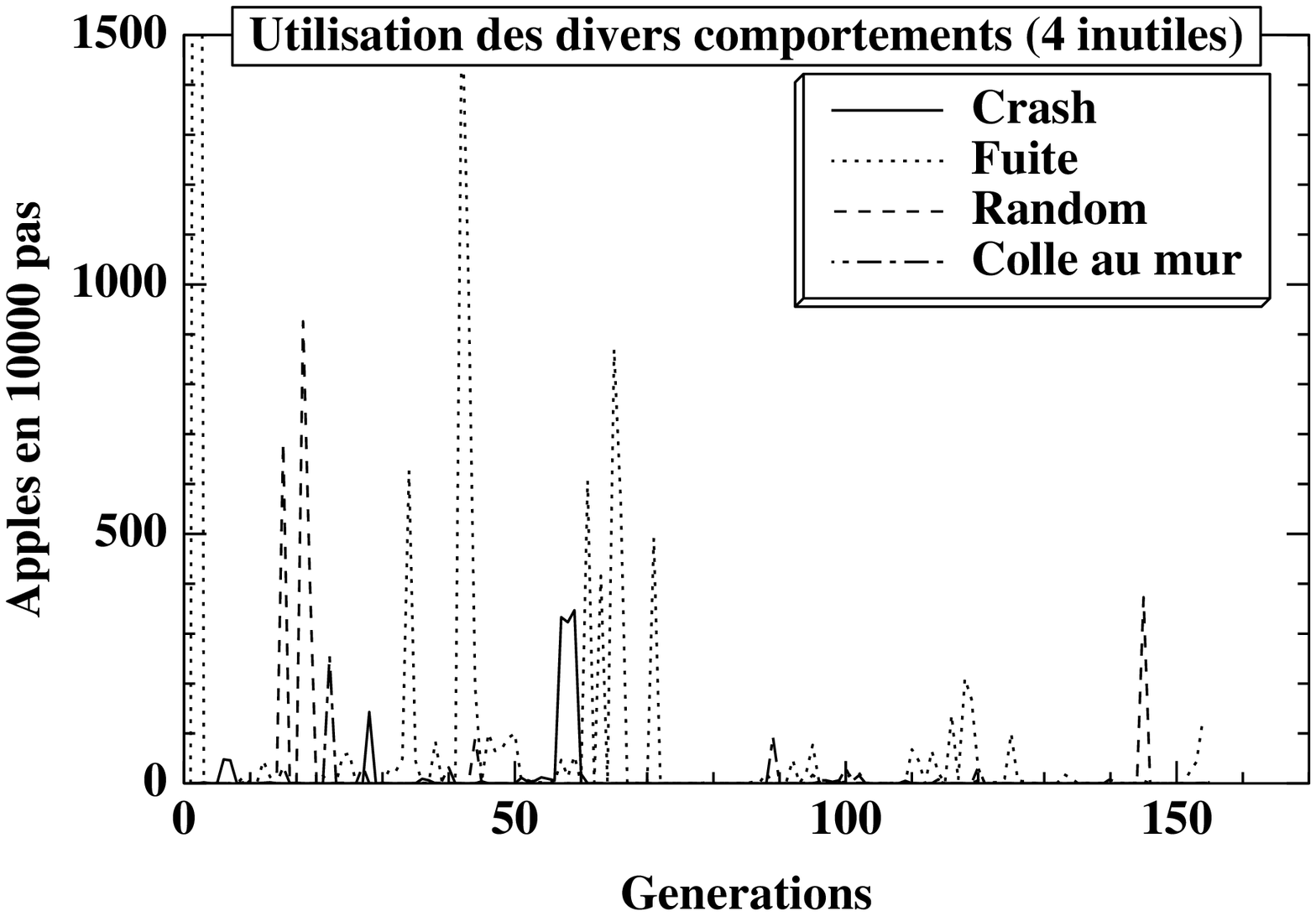}
\end{tabular}

    \caption{Number of calls out of 10000 time steps of the 
       {\bf useful} (left) and {\bf useless} (right) behaviors.
The plots for the useful behaviors are roughly the same whether or not
some useless          behaviors are available.}
    \label{behaviorcalls}
\end{figure}
\end{center}

First, 4 behaviors were added to the existing ones:  {\em random},
{\em light avoiding}, {\em crash} (goes straight into the nearest
wall!) and {\em stick to the walls} (tries to stay close to a wall). 
The first generations demonstrated lower performances and higher
variance than in the
initial experiment, as all behaviors were used with equal
probability. However, the plot of the best fitness (not shown) soon
catches up with the plots of Figure \ref{bestavg4}, and after 150
generations, the results are hardly distinguishable. Looking at the
frequency of use of each behavior, it clearly appears that the same
useful behaviors are used (see Figure \ref{behaviorcalls}-left, and section
  \ref{analysis} for a discussion). Moreover, the useless behaviors
  are scarcely used as evolution goes along, as can be seen on Figure
  \ref{behaviorcalls}-right (beware of the different scale).

These good stability results have been confirmed by adding 20 useless
behaviors (5 times the same useless 4). The results are very similar -
though a little worse, of course, as the useless behaviors are called
altogether a little more often.

\vspace{-1cm}

\subsection{Generalization}
\label{generalization}
Several other experiments were performed in order to test the
generalization abilities of the resulting controllers.  The 
$10 \times 4$ best 
controllers  obtained in the experiments above
were tested in new experimental environments. 

First, some obstacles were added in the environment (the arena was
free of obstacle in the experiments described above). But no
controller was able to go to the recharge area whenever an obstacle
was in the way. However, when the evolution was performed with the
obstacles, the overall results are about the same (with slightly worse
overall performance, as predicted).
\revise{Some experiments were also run with a moving recharge area, without
any significant modification of the results.}{~}

More interesting are the results obtained when the robots are put in
an arena that is three times larger than the one used during
evolution. The best generalization results are obtained by the
Classical Controller architecture, with only a slight decrease of
fitness(a few percents). Moreover, 100 additional generations of
evolution in the new environment gives back the same level of fitness.

The Symbolic Controllers come next (in terms of generalization
capability!): they initially lose around 10\% of their fitness, then
reach the same level again in 150 generations, and even overpass that
level (see the discussion in section \ref{complexresults}).

Surprisingly, both supervisor architectures fail to reach the same
level of performance in the new environment even after a new cycle of
evolution. The Symbolic Supervisor lose about 12.5\% of their fitness,
and only recover half of it, while the Classical Supervisors lose more
than 20\% and never recover. 

These results can be at least partly explained by the behaviors that
are obtained by the first experiments: whereas all direct controller
architectures mainly stay around the recharge area, and thus are not
heavily disturbed by the change of size of the arena, the Supervisor
architectures use their exploration behavior and fail to turn back on
time. The only surprising result is that they also fail to reach the
same level of fitness even after some more generations
\footnote{However, when restarting the evolution from scratch in the
  large arena, the SSs easily reach the $6400$ fitness level,
  outperforming again all other architectures}.

This difference in the resulting behaviors also explains the results
obtained in the last generalization experiment that will be presented
here: the recharge of energy was made 2 times slower (or the energy
consumption was made twice faster -- both experiments give exactly the
same results).
Here, the results of the Symbolic Supervisors are clearly much
better than those of the other architectures: in all cases, the robot
simply stays in the recharge area {\bf until the energy level is back
  to maximum}, using the {\em stop} behavior.

Surprisingly, most Classical Supervisors, though they also can use their
{\em STOP} behavior, fail to actually reach the recharge area.
On the other hand, both direct controller architecture never stop on
the recharge area. However, while the Symbolic Controllers manage to
survive more than one epoch for half of the trials, all Classical
Controllers fail to do so.

This last generalization experiment shows a clear advantage to the
Symbolic Controller architecture: if is the only one that actually
learned to recharge the accumulator to its maximum before leaving the
recharge area.
\revise{~}{
But the ultimate test for controllers evolved using a simulator is of
course to be applied on the real robot. This is on-going work, and
the first experiments, applied to the obstacle avoidance behaviors,
have confirmed the good performance of the symbolic controllers in
any environment. 
}

\section{Discussion and Perspectives}
\label{conclusion}
The main contribution of this work is to propose some compromise
between the pure black box approach where evolution is supposed to
evolve everything from scratch, and the ``transparent box'' approach,
where the programmer must decompose the task manually.

The proposed approach is based on a toolbox, or library, of behaviors
ranging from 
elementary hand-coded behaviors to evolved behaviors of low to medium
complexity. The combination and proper use of those tools is left to
evolution. The new space of controllers that is explored is more
powerful that the one that is classically explored in Evolutionary
Robotics. For instance, it was able to easily find some loophole in the (very
simple) obstacle behavior fitness; moreover, it actually discovered
the right way to recharge its accumulator in the more complex homing
experiment.

Adding new basic behaviors to that library allows one to gradually
increase the complexity of the available controllers without having to
cleverly insert those new possibilities in the available controllers:
evolution will take care of that, and the sensitivity analysis
demonstrated that useless behaviors will be filtered out at almost no
cost (section \ref{sensitivity}). For instance, there might be some
cases where a random behavior can be beneficial 
-- and it didn't harm the energy experiment.
More generally, this idea of a library allows one to store experience from
past experiments: any controller (evolved or hand-coded) can be added
to the toolbox, and eventually used later on - knowing that useless
tools will simply not be used.

Finally, using such a library increases the intelligibility of the
resulting controllers, and should impact the way we evolutionary design
controllers, i.e. fitness functions. One can add some
constraints on the distribution over the use of the different available
controllers, (e.g. use the light following action $\varepsilon$\% of
the time); by contrast, traditional evolutionary approach had to set
up sophisticated {\em ad hoc} experimental protocol to reach the same
result (as in \cite{Tuci:Harvey:SAB02}). Further work will have to
investigate in that direction.

But first, more experiments are needed to validate the proposed
approach (e.g. experiments requiring some sort of memory, as in
\cite{Yamauchi:Beer:SAB94,Tuci:Harvey:SAB02}). The impact
of redundancy will also be investigated: in many Machine Learning tasks,
adding redundancy improves the quality and/or the robustness of the
result. Several controllers that have been evolved for the same task,
but exhibit different behaviors, can be put in the toolbox. 
It can also be useful to allow the overall controller to use all
levels of behaviors simultaneously instead of the layered architecture
proposed so far. This should allow to discover on the fly specific
behaviors 
whenever the designer fails to include them in the library.

Alternatives for the overall architecture will also be looked for. One
crucial issue in autonomous robotics is the adaptivity of the
controller. Several architectures have been proposed in that direction
(see \cite{Nolfi:Floreano:Book00} and references herein) and will be
tried, like for instance the idea of auto-teaching networks.

Finally, in the longer run, the library approach helps to keep tracks
of the behavior of the robot at a level of generality that can be
later exploited by some data mining technique. Gathering the Frequent
Item Sets in the best evolved controllers can help deriving some
brand new macro-actions. The issue will then be to check how
useful such macro-actions can be if added to the library.

\vspace{-0.3cm}



\end{document}